\documentclass[conference]{IEEEtran}
\IEEEoverridecommandlockouts
\usepackage{cite}
\usepackage{bm}

\usepackage{amsmath,amssymb,amsfonts}
\usepackage{algorithmic}
\usepackage{graphicx}
\usepackage{textcomp}
\usepackage{booktabs}
\usepackage{url}
\usepackage{xcolor}
\def\BibTeX{{\rm B\kern-.05em{\sc i\kern-.025em b}\kern-.08em
    T\kern-.1667em\lower.7ex\hbox{E}\kern-.125emX}}
\begin{document}

\title{Frequency maps reveal the correlation between Adversarial Attacks and Implicit Bias
}
\author{
	\IEEEauthorblockN{Lorenzo Basile$^1$, Nikos Karantzas$^2$, Alberto d'Onofrio$^1$, Luca Manzoni$^1$, \\Luca Bortolussi$^1$, Alex Rodriguez$^{1,3,\dagger}$, Fabio Anselmi$^{1,4,\dagger}$}\\
	\IEEEauthorblockA{$^1$University of Trieste, Italy, $^2$Stanford University, USA, $^3$ICTP, Italy, $^4$MIT, USA}
	\IEEEauthorblockA{$^\dagger$Equal supervision}
	\IEEEauthorblockA{Correspondence to: \texttt{fabio.anselmi@units.it}}
}

\maketitle

\begin{abstract}
Despite their impressive performance in classification tasks, neural networks are known to be vulnerable to adversarial attacks, subtle perturbations of the input data designed to deceive the model. 
In this work, we investigate the correlation between these perturbations and the implicit bias of neural networks trained with gradient-based algorithms. 
To this end, we analyse a representation of the network's implicit bias through the lens of the Fourier transform.
Specifically, we identify unique fingerprints of implicit bias and adversarial attacks by calculating the minimal, essential frequencies needed for accurate classification of each image, as well as the frequencies that drive misclassification in its adversarially perturbed counterpart. This approach enables us to uncover and analyse the correlation between these essential frequencies, providing a precise map of how the network’s biases align or contrast with the frequency components exploited by adversarial attacks.
To this end, among other methods, we use a newly introduced technique capable of detecting nonlinear correlations between high-dimensional datasets. 
Our results provide empirical evidence that the network bias in Fourier space and the target frequencies of adversarial attacks are highly correlated and suggest new potential strategies for adversarial defence. Code is available at \url{https://github.com/lorenzobasile/ImplicitBiasAdversarial}
\end{abstract}


\section{Introduction}
Regardless of their overparameterisation, deep neural networks can avoid over-fitting and generalise well to new data. This surprising property has been attributed to the so-called implicit bias of the network, which refers to the ability of gradient descent to converge to solutions with good generalisation properties among the many that correctly label the training data \cite{li2019enhanced,zhang2017understanding,arora2019exact}.
Understanding this bias has been the focus of extensive research in recent years. However, it remains unclear how it leads to generalisation in nonlinear, complex neural networks.
Indeed, with the exception of simple models \cite{gunasekar2018implicit}, a formal characterisation of the implicit bias of a neural network remains a formidable challenge.

Besides generalisation, another important (and related) property of neural networks is robustness, i.e., their ability to maintain performance and accuracy when exposed to noise in input data and adversarial attacks.  
In particular, adversarial attacks have gained significant attention since the work of \cite{szegedy2013intriguing}, as they consist of perturbations of the input that are imperceptible to humans but can drastically alter the response of a trained network. 
Although substantial research has been conducted to understand this vulnerability \cite{goodfellow2015explaining,athalye2018obfuscated,ilyas2019adversarial,wu2020making} and numerous strategies have been developed to mitigate its effects \cite{papernot2016distillation,kurakin2017adversarial,madry2018towards,wong2020fast}, a definitive explanation of its origins remains elusive. 

In this work, we explore the relationship between two crucial properties of a trained network: its vulnerability to adversarial attacks and its implicit bias. While this problem has been the focus of recent studies, explicit results have so far only been achieved for simple networks \cite{frei2023benign,frei2023implicit,vardi2022implicit}. In this context, recent research \cite{karantzas2022understanding} has developed algorithms to investigate specific aspects of implicit bias within complex networks, particularly by analysing the frequencies that are essential for accurate classifications. This approach involves training a modulatory mask for each input image, which selectively filters frequency content and reduces it to the bare minimum needed to maintain the correct classification. In doing so, trained masks encode a map of the essential frequencies required by the network. Hence, in the rest of this manuscript we will refer to them as frequency maps. In our work, we employ such frequency maps to explore, for each input, the relationship between the critical frequencies needed for accurate classification and those targeted by adversarial attacks for misclassification. These maps serve as a dual representation: they reveal the network’s implicit bias in Fourier space by highlighting the frequencies it relies on to process inputs, while also illustrating the specific frequency components used by adversarial attacks.

\begin{figure*}[t]
  \centering
  \includegraphics[width=0.95\textwidth]{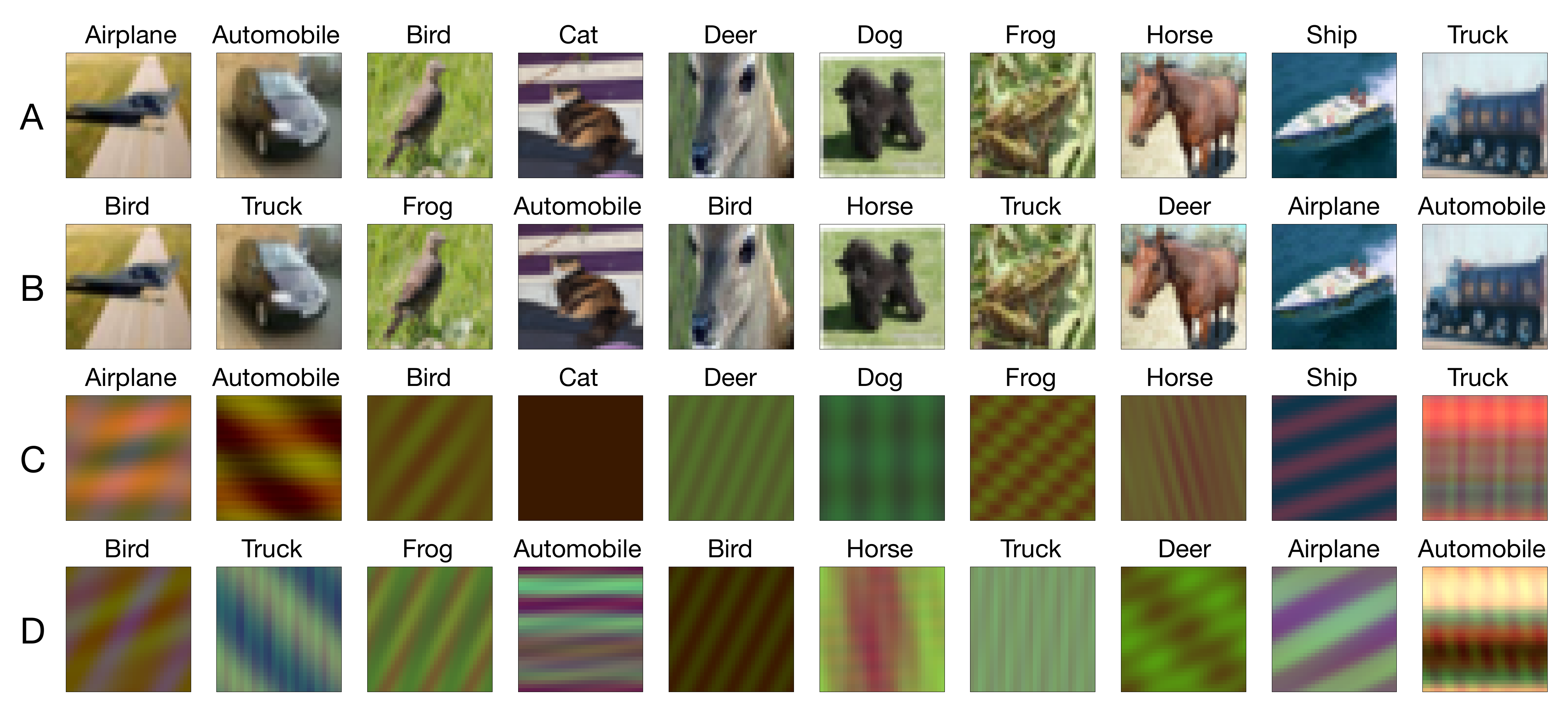}
  \caption{Examples of CIFAR-10 \cite{krizhevsky2009learning} images before and after being filtered by the Fourier maps: (A): original input images (B): adversarial images generated with $\ell_\infty$ Fast Minimum Norm \cite{pintor2021fast} attack on ResNet-20 \cite{he2016deep} (C): images filtered by essential frequency maps (D): adversarial images filtered by adversarial frequency maps.}\label{allmaps}
\end{figure*}

Fig.~\ref{allmaps} displays examples of clean and attacked images before (A, B) and after (C, D) being filtered by, respectively, essential frequency maps and adversarial frequency maps.
Our primary objective is to provide empirical proof that the network bias is correlated with the adversarial attacks in Fourier space.
However, defining and computing this correlation is challenging due to the high-dimensional nature of the modulatory map sets and their potential highly nonlinear correlation.
To address these challenges we use various correlation estimators, among which a recently introduced nonlinear one \cite{basile2025intrinsic}, that relies on the observation that the intrinsic dimensionality ($I_d$) of a dataset is closely linked to the correlations among the various features defining the data points.
Our findings indicate a strong correlation between the manifolds of the feature spaces defined by the two types of maps, providing empirical evidence of the similarity between the network bias in Fourier space and the target frequencies of adversarial attacks.
The main contributions of this work can be summarised as follows:
\begin{itemize}
\item We provide the first detailed quantitative computation of adversarial essential frequencies, which constitute a representation of the features targeted by adversarial attacks. 
\item We demonstrate a strong correlation between the network's implicit bias and the adversarial perturbations through the essential frequencies of clean and adversarial images.
\item We show that the essential frequencies required to classify images in a certain class are sufficient to counteract adversarial attacks, suggesting a promising approach for designing new defence protocols.
\end{itemize}

\section{Related work and background}

\subsection{Implicit Bias, Implicit Fourier Bias and Robustness}
\begin{figure*}[t!]
  \centering
  \includegraphics[width=0.9\textwidth]{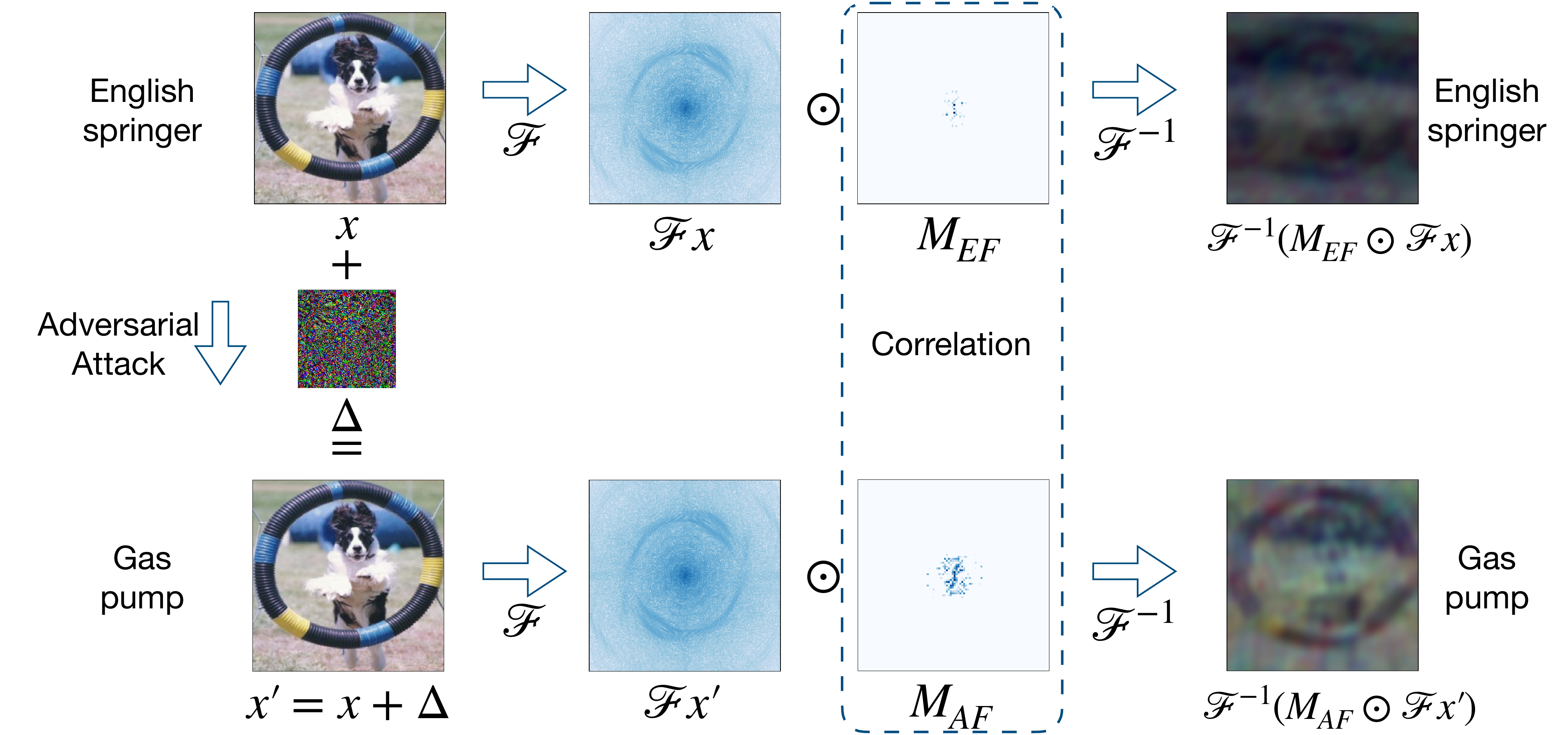}
  \caption{Schematic representation of the method employed to obtain essential frequency maps and adversarial frequency maps. Only one channel is displayed for visualization purposes. Full details are provided in Sec.~\ref{map_training}.}\label{method}
\end{figure*}
Overparameterised networks are characterised by a loss landscape with numerous local minima that correctly label the training data.
The idea behind the phenomenon of implicit bias 
is that which local minimum the model converges to after training depends on the complex interplay between different factors including the choice of the architecture \cite{gunasekar2018implicit, yun2020unifying}, the weights initialisation scheme \cite{sahs2022shallow}, the optimisation algorithm \cite{williams2019gradient, woodworth2020kernel} and the data statistics \cite{yin2019fourier}.

The implicit bias of state-of-the-art models has been shown to play a critical role in the generalisation properties of deep neural networks \cite{li2019enhanced,arora2019exact}. 
Research over the years has thoroughly investigated the implicit bias present in training algorithms for different network architectures. 
Although an explicit characterisation of the bias has been found only in simple networks such as linear \cite{saxe2014exact,gunasekar2017implicit,woodworth2020kernel,min2021explicit}, and some nonlinear models \cite{boursier2022gradient,wang2023understanding,frei2023implicit,frei2023the,abbe2023transformers}, the phenomenon is still lacking a final explanation.
Interestingly, a study by Frei et al. \cite{frei2023implicit} demonstrates that the implicit bias in neural networks acts as a "Double-Edged Sword": while it enhances generalisation properties, it can also compromise the network's robustness, presenting both benefits and potential drawbacks. Similar conclusions are reached in \cite{tsilivis2024price}. 
Conversely, \cite{faghri2021bridging} reveals a deep connection between robust classification and implicit bias, suggesting that the implicit bias can improve robustness in linear regression.
Finally, \cite{min2024can} show that a shallow network with a polynomial ReLU activation trained by gradient flow not only generalises well but is also robust to adversarial attacks.
These studies highlight that, while some insight into the relationship between implicit bias and the robustness of a neural network has been gained, a clear understanding is still lacking.
One interesting effect of the implicit bias in neural networks is their tendency to learn frequencies from low to high in the \textit{target function} during training, a phenomenon known as spectral bias \cite{rahaman2019spectral}. This bias leads the network to learn low-complexity functions, potentially explaining its ability to generalise \cite{fridovich2022spectral, cao2021towards, wang2020high, tsuzuku2019structural}. Additionally, the implicit bias can also significantly influence the types of input features extracted by a trained neural network. For instance, \cite{karantzas2022understanding} demonstrate that very few image frequencies in the Fourier domain are enough for the network to perform classification. These findings have helped characterise the spectral bias of neural networks with a focus on the input space rather than the target function, as in \cite{rahaman2019spectral}.
Moreover, empirical evidence shows a strong relationship between network robustness and the statistics of the Fourier spectra of the input data \cite{yin2019fourier} or architecture \cite{caro2024translational}. In particular, certain frequency-based patterns in adversarial robustness have been identified, suggesting that low-frequency perturbations can be especially effective \cite{sharma2019effectiveness}. Furthermore, studies indicate that frequency-driven perturbations can be leveraged for adversarial attacks while remaining imperceptible to human observers \cite{luo2022frequency}. Detection strategies in the Fourier domain have also been used to defend against adversarial attacks \cite{harder2021spectraldefense}.

\subsection{Adversarial Attacks}

Artificial Neural Networks are notoriously vulnerable to adversarial attacks \cite{szegedy2013intriguing, goodfellow2015explaining}.
These attacks involve manipulating an input data point in a way that deceives an otherwise well-performing classifier, by making small alterations to a correctly classified data point. Numerous techniques have been proposed to create such adversarial examples, beginning with the Fast Gradient Sign Method (FGSM) \cite{goodfellow2015explaining}, followed shortly by variants such as Projected Gradient Descent (PGD) \cite{madry2018towards}. Both these methods employ gradient information to generate an appropriate adversarial example while ensuring that the $\ell_p$ norm of the perturbation remains below a fixed threshold $\epsilon$. 
These algorithms were primarily developed for effectiveness rather than optimality, which may limit their ability to generate input samples with minimal perturbations, resulting in them being classified as "maximum confidence" attacks. In contrast, "minimum norm" attacks prioritise the identification of adversarial examples with the least amount of perturbation by minimising its norm.
In this regard, some of the most notable proposals are L-BFGS \cite{szegedy2013intriguing},  the Carlini and Wagner attack \cite{carlini2017towards}, DeepFool \cite{moosavi2016deepfool} and the more recent Fast Minimum Norm (FMN) attack~\cite{pintor2021fast}, which seeks to combine the efficiency of FGSM and PGD with optimality in terms of perturbation norm.

The robustness of neural networks against adversarial attacks remains an unresolved issue. 
Although adversarial training is currently the most effective technique for improving the resilience of neural classifiers, it often involves a trade-off between robustness and a reduction in performance on non-adversarial, clean data \cite{goodfellow2015explaining, raghunathan2020understanding}. Moreover, it remains unclear why adversarial examples exist and whether they represent an inevitable byproduct of current neural architectures and training methods \cite{ilyas2019adversarial, shafahi2018adversarial}. The goal of this work is not to propose a method for improving the adversarial robustness of neural networks. Rather, our aim is to provide valuable insights into the frequency content that is targeted by adversarial attacks and its relationship with the implicit spectral bias of the network.

\section{Methods}
\subsection{Modulatory Maps}
\label{maps}

Our main tools for understanding the implicit spectral bias and the geometry of adversarial examples are modulatory maps. These maps are computed to extract information about the essential input frequencies that the network requires to perform a specific classification task.
To generate the maps, we employ a procedure similar to the one described in \cite{karantzas2022understanding}, as illustrated in Fig.~\ref{method}. We train maps to modulate the frequency content of an image by performing element-wise multiplication between each entry of the Fast Fourier Transform (FFT) of the image and the corresponding entry of the map. Each entry of the map is a learnable scalar between $0$ and $1$.
Specifically, starting from an image $x$, we compute its FFT $\mathcal{F}x$ and multiply it element-wise with a learnable map $M$. The map has the same shape of the image $x$ (and its FFT $\mathcal{F}x$), meaning that if the image has RGB encoding we train a separate map for each channel, and its entries are constrained to be in $[0,1]$. The result of this multiplication is then projected back in pixel space by taking the real part of its inverse Fourier transform, thereby obtaining a new filtered image $x_F$: 

\begin{equation}
x_F=\Re({\mathcal{F}^{-1}(M\odot \mathcal{F}x)}).
\end{equation}

\begin{figure}[h]
  \centering
  \includegraphics[width=\linewidth]{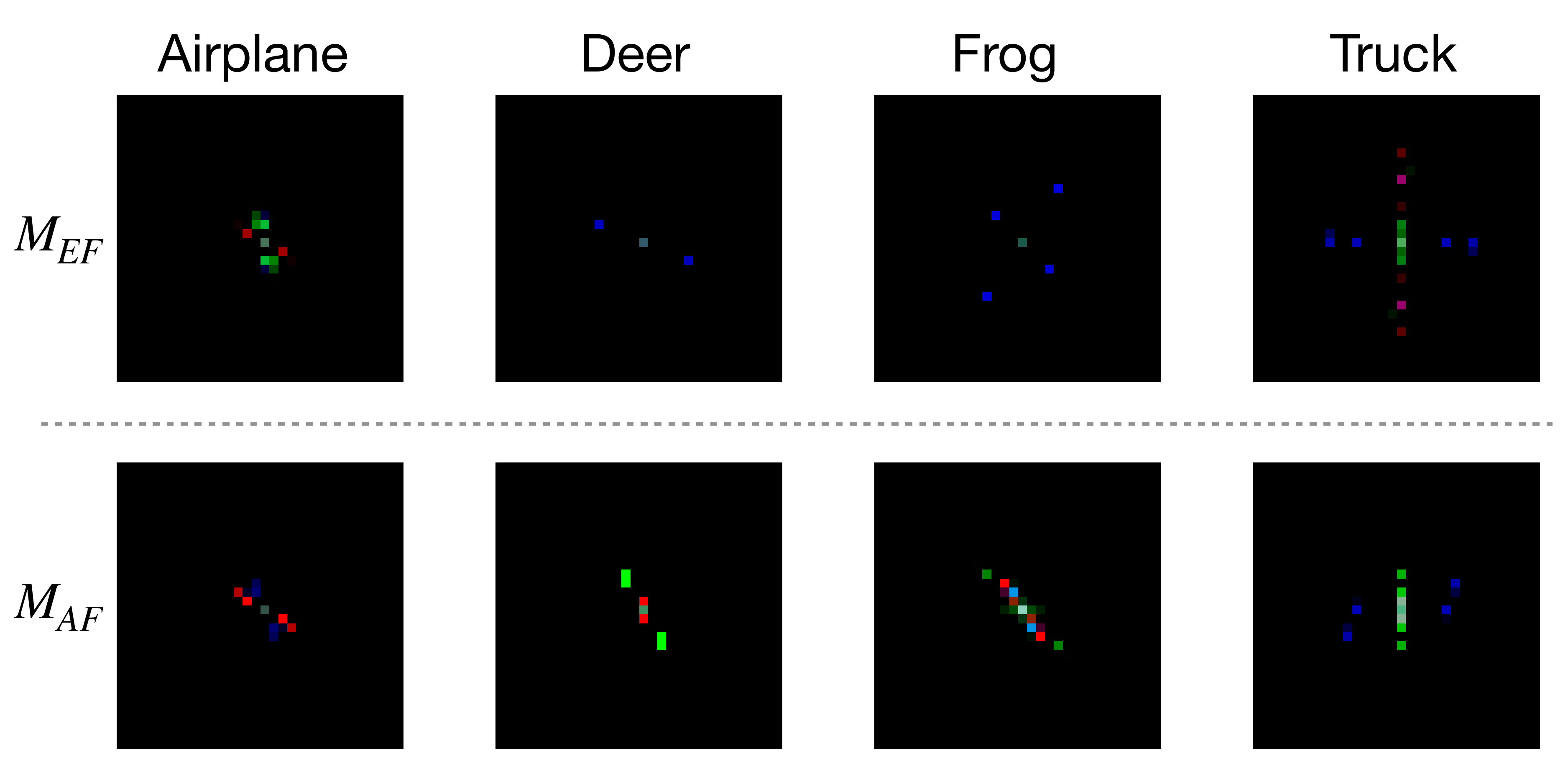}
  \caption{Examples of essential and adversarial frequency maps, represented as RGB images. The labels refer to the classification of the clean image. The maps were obtained using CIFAR-10 and the Fast Minimum Norm attack on ResNet-20.}\label{adversarial_maps}
\end{figure}

The image $x_F$ is then fed into the trained classification model to obtain a prediction. We produce two sets of maps. The maps belonging to the first set encode the essential frequencies of an image to be correctly classified by the neural classifier, thus we will refer to these as \textit{essential frequency maps} ($M_{EF}$). The second set is composed of maps that encode the essential frequency content required to maintain the effectiveness of an adversarial attack, that is, the essential frequencies needed to misclassify an adversarially perturbed image. We will refer to these maps as \textit{adversarial frequency maps} ($M_{AF}$). Some examples of essential and adversarial frequency maps are shown in Fig. \ref{adversarial_maps}. Both sets of maps are learned using a preprocessing layer attached to a trained classifier with frozen parameters. 
The essential frequency maps are trained by optimising the Cross-Entropy loss of the entire model (consisting of the preprocessing layer and the trained classifier) on the original samples. Conversely, for adversarial frequency maps, the training objective is the Cross-Entropy with respect to the adversarial class (to preserve misclassification), and the maps are trained on adversarial data.
The key property of the learned maps is their \textit{sparsity}, which is achieved by enforcing an $\ell_1$ norm regularisation on the entries of the map during training. This regularisation ensures that the map accurately captures only the essential frequency content needed to accomplish a specific task, such as correctly classifying an input or misclassifying an adversarial example.
Our primary objective is to generate these maps and to determine whether a correlation exists between these distinct sets of maps.

\subsection{High-dimensional correlation}\label{corr_methods} 
As mentioned earlier, to examine the statistical relationship between implicit bias and adversarial attacks, we compute the correlations between two feature spaces: that of the essential frequencies for image classification and that of the essential frequencies required for the adversarial attack to be successful. In this section we introduce the main tools that we employ to this aim. 

The most common approach for investigating correlations is based on the Pearson correlation coefficient ($R^2$) between variables. However, this method can only be applied to univariate datasets, making it impractical when dealing with multidimensional data, like in our case. The direct analogous to $R^2$ in a multivariate scenario is Canonical Correlation Analysis (CCA) \cite{hotelling1936relations}. CCA works by linearly projecting the two multidimensional datasets to a subspace where the correlation between the projected variables is maximised. A more modern derivative of CCA is Singular Value CCA (SVCCA) \cite{raghu2017svcca}, which employs Principal Component Analysis before CCA, with the objective of denoising and reducing the dimensionality of data.
SVCCA proved to be particularly useful in analysing the internal representations of neural networks \cite{raghu2017svcca}. 

In this direction, other correlation metrics have been proposed, including Projection Weighted CCA \cite{morcos2018insights}, Centered Kernel Alignment \cite{kornblith2019similarity} and the recent Intrinsic Dimension Correlation ($I_d$Cor) \cite{basile2025intrinsic}. $I_d$Cor is a method for quantifying correlation that is well suited for nonlinear multidimensional spaces. It is based on the concept of Intrinsic Dimension ($I_d$) of data, which refers to the minimum number of variables required to describe a dataset. The key idea behind $I_d$Cor is that if there is a correlation between two datasets (i.e., the information from one can partially reconstruct the other), the combined $I_d$ of the datasets will be lower than the sum of their individual $I_d$s. This method provides both a correlation coefficient and a P-value, obtained through a permutation test, indicating the probability of no correlation.

Finally, Cosine Similarity is another method for measuring similarity between high-dimensional vectors, defined as the cosine of the angle between two vectors. Unlike other methods, cosine similarity is not strictly a measure of correlation. It offers a localised measure of similarity for individual pairs of vectors rather than evaluating entire datasets (i.e., multiple samples of a random variable). Additionally, cosine similarity requires that both vectors exist in the same vector space, making it unsuitable for vectors with different dimensionalities.

\section{Experimental results}\label{results}

\subsection{Data}
\looseness=-1 For our experiments, we employ two benchmark image datasets: CIFAR-10 \cite{krizhevsky2009learning} and Imagenette \cite{imagenette}. CIFAR-10 consists of $60000$ RGB $32\times32$ training images and $10000$ test images categorised into $10$ classes. When studying adversarial examples, we use the test images, and the training set is solely employed for fine-tuning models, as explained in greater detail in the subsequent section.
Imagenette, a $10$-class subset of ImageNet \cite{russakovsky2015imagenet} is used to scale up our experiments to a higher-dimensional data set. In this case, we resize images to $224\times224$ and use both training and test images (respectively, 9469 and 3925 images) to train modulatory maps, as detailed in Sec. \ref{imagenet}.
\subsection{Models}
To gain a more complete picture of how implicit bias and adversarial attacks are related in various scenarios, we employ classification models based on different neural architectures.
For CIFAR-10 data, we use ResNet-20, a relatively small representative of the very well known ResNet family, introduced in \cite{he2016deep}. We train the model from scratch, and provide details on the training procedure in the Appendix (Sec.~\ref{train_details}).
In the case of Imagenette, we employ both a CNN (ResNet-18 \cite{he2016deep}) and a Vision Transformer (ViT-B/16 \cite{dosovitskiy2020image}). We take these models pretrained on ImageNet-1K and we only train the final classification head, to adapt it to the classification task at hand.
\subsection{Attacks}
\looseness=-1 We employ three $\ell_\infty$ gradient-based adversarial attacks: the Fast Minimum Norm (FMN) attack \cite{pintor2021fast}, Projected Gradient Descent (PGD) \cite{madry2018towards} and DeepFool \cite{moosavi2016deepfool}. All the attacks are considered in the untargeted setting. While FMN and DeepFool do not depend on any hyperparameter, for PGD it is necessary to allocate a perturbation budget $\epsilon$. For most experiments, we set it to the standard value of $\epsilon=\frac{8}{255}$ \cite{croce2021robustbench}, but we provide an analysis of the robustness of our findings with respect to $\epsilon$ in Sec. \ref{variable_eps}.
To implement all attack algorithms, we use the implementation provided in the Foolbox library \cite{rauber2017foolbox, rauber2017foolboxnative}. 

\subsection{Map training}
\label{map_training}
\looseness=-1 The key step in our experimental procedure is the training of Fourier maps (see Sec.~\ref{maps}). Starting from a trained, well-performing classifier, we freeze its parameters and prepend to it a pre-processing layer that computes the FFT of an image, multiplies it element-wise by the trainable map and computes the inverse FFT. The real part of the resulting image is then fed into the classifier. The process of training the maps is identical for both sets of modulatory maps, with the only difference being the data set used for map training. We train essential frequency maps using clean images associated with their original labels. In contrast, for adversarial frequency maps, we utilise adversarial images and the adversarial labels produced by the classifier for those images. In this step, we optimise the standard Cross-Entropy loss function with the addition of an $\ell_1$ penalty term to promote map sparsity. Further details on the map training procedure are reported in Sec.~\ref{map_training_app} in the Appendix.

\subsection{Correlation between adversarial and essential frequency maps}
\subsubsection{CIFAR-10}
To provide evidence of the relation between the implicit bias of the network and the adversarial perturbations, we adopt a direct approach: we correlate the essential frequency maps with the adversarial frequency maps. 
This correlation analysis is performed using various methods, introduced in Sec. \ref{corr_methods}. Namely, we employ SVCCA \cite{raghu2017svcca} as a linear baseline, Cosine Similarity, and Intrinsic Dimension Correlation ($I_d$Cor \cite{basile2025intrinsic}).

Correlation results for ResNet-20 on CIFAR-10 data, reported in Table \ref{correlations_1}, show that there is a clear correlation between the essential frequencies of the classifier and those of adversarial attacks. While linear correlation, measured by SVCCA, is not strikingly high, $I_d$Cor finds consistent evidence of correlation, as witnessed by the correlation scores above $0.6$ for all attacks and high confidence ($\text{P-value}=0.01$). Interestingly, even a very simple metric such as cosine similarity finds strong similarities between the two sets of maps, as average similarity is just below $0.5$ for all attacks. However, cosine similarity is a local measure, returning one value for each pair of maps, and high average similarity comes at the cost of extremely high fluctuations among individual data points, as standard deviations exceed $0.25$ for all attacks.

\looseness=-1 Summing up, these results point to a strong correlation between the essential and adversarial frequencies of the classifier. This correlation has a significant nonlinear component, as witnessed by the advantage shown by $I_d$Cor with respect to SVCCA. Moreover, its nature is strongly impacted by the individual features of images (as shown by high variations in cosine similarities).
\begin{table}[h]
\centering
  \caption{Correlation between essential frequency maps and adversarial frequency maps (CIFAR-10, ResNet-20). (\textbf{SVCCA}): SVCCA \cite{raghu2017svcca} correlation coefficient between essential and adversarial frequency maps; (\textbf{Cosine sim}) mean $\pm$ std. dev. cosine similarity between maps; ($\mathbf{I_d}$\textbf{Cor}) correlation coefficient between maps, according to $I_d$Cor \cite{basile2025intrinsic}; ($\mathbf{P}$\textbf{-value}) P-value for $I_d$Cor (low means high probability of correlation)}
  \label{correlations_1}
  \begin{tabular}{ccccc}
    \toprule
    \textbf{Attack} & \textbf{SVCCA} & \textbf{Cosine sim} & $\mathbf{I_d}$\textbf{Cor} & $\mathbf{P}$\textbf{-value}\\
    \midrule
    FMN       & $0.25$ & $0.49 \pm 0.27$ & $0.66$ & $0.01$ \\
    PGD       & $0.30$ & $0.49 \pm 0.28$ & $0.60$ & $0.01$ \\
    DeepFool  & $0.25$ & $0.48 \pm 0.27$ & $0.62$ & $0.01$ \\
    \bottomrule
  \end{tabular}
\end{table}

\subsubsection{Imagenette}\label{imagenet}

We now move to a larger-scale experiment using Imagenette \cite{imagenette}, a 10-class subset of ImageNet \cite{russakovsky2015imagenet}. The training of modulatory maps (essential frequency maps and adversarial frequency maps) is conducted following the same procedure used for CIFAR-10, with the only difference being that we use both the training and test images to calculate the maps. This choice is due to the reduced size of Imagenette's test set, which could impact the accuracy of subsequent analyses. Indeed, an accurate estimation of intrinsic dimension is crucial when using $I_d$Cor \cite{basile2025intrinsic} as the number of data points needed for reliable estimation scales exponentially with the intrinsic dimension itself \cite{bac2021scikit}.
For Imagenette, we compute the same correlation statistics as provided for CIFAR-10 (SVCCA, $I_d$Cor, and cosine similarities). However, due to the high resolution of maps (three channels of size $224 \times 224$), repeated intrinsic dimension estimations are computationally infeasible on raw data. Therefore, we average-pool maps to $32 \times 32$ pixels before applying correlation methods.
The results are reported in Table \ref{correlations_imagenet}. Similar to the previous dataset, the discrepancy between $I_d$Cor and SVCCA reveals the presence of a nonlinear correlation. Correlation is found for all combinations of adversarial attack (FMN, PGD, and DeepFool) and model architecture (ResNet and ViT).

\begin{table}[h]
\centering
  \caption{Correlation between essential frequency maps and adversarial frequency maps (Imagenette).\\
  }
  \label{correlations_imagenet}
  \centering
  
  \begin{tabular}{ccccccc}
    \toprule
    \textbf{Attack} & \textbf{Model} & \textbf{SVCCA} & \textbf{Cosine sim} & $\mathbf{I_d}$\textbf{Cor} & $\mathbf{P}$\textbf{-value}\\
    
    \midrule
    FMN & ResNet-18 & $0.23$ & $0.60\pm 0.15$ & $0.63$ & $0.01$ \\
     & ViT-B/16 & $0.21$ & $0.56\pm 0.18$ & $0.59$ & $0.01$ \\
    PGD & ResNet-18  & $0.19$ & $0.60\pm 0.16$ & $0.62$ & $0.01$ \\
     & ViT-B/16  & $0.23$ & $0.53\pm 0.20$ & $0.58$ & $0.01$ \\
    DeepFool & ResNet-18  & $0.23$ & $0.59\pm 0.15$ & $0.62$ & $0.01$ \\
    & ViT-B/16  & $0.22$ & $0.53\pm 0.19$ & $0.56$ & $0.01$ \\
    \bottomrule
  \end{tabular}
\end{table}

\subsection{Correlation with variable perturbation magnitude}\label{variable_eps}

The PGD attack algorithm allows to tune the parameter $\epsilon$, which controls the magnitude of the adversarial perturbation in terms of $\ell_\infty$ norm. Throughout the paper, we kept this value fixed at $\frac{8}{255}$. However, varying this parameter can provide useful insight into our results in terms of correlation between essential frequency maps and adversarial frequency maps. Interestingly, as shown in Table \ref{var_eps_tab}, the correlation does not directly depend on the magnitude of the adversarial perturbation, and it emerges with high confidence even with more subtle attacks, with $\epsilon=\frac{1}{255}$ and $\epsilon=\frac{3}{255}$.

\begin{table}[ht]
\centering

\caption{Attack success rates and $I_d$Cor correlation results for PGD with variable perturbation magnitude $\epsilon$ on ResNet-20.}
\label{var_eps_tab}
\centering
\begin{tabular}{ccccccc}
    \toprule
    \textbf{$\bm{\epsilon}$} & \textbf{Attack success rate} & \textbf{$I_d$Cor} & \textbf{P-value}
    \\
    \midrule
    $\frac{1}{255}$ & $68.56\%$ & $0.59$ & $0.02$ 
    \\
    $\frac{3}{255}$ & $99.78\%$ & $0.53$ & $0.04$ 
    \\
    $\frac{8}{255}$ & $100.00\%$ & $0.60$ & $0.01$
    \\
    \bottomrule

\end{tabular}
\end{table}

\subsection{Modulatory maps can revert adversarial attacks}
\label{robust_maps}

We now move to assessing whether essential frequency maps can help mitigate the effect of adversarial attacks. The first step we take in this direction is overcoming the one-to-one relationship between input images and frequency maps by training a single map that encodes the essential frequency content for an entire class. Such class-level essential frequency maps can be obtained by following the same approach used to learn essential frequency maps for single images, but training on all the images belonging to a certain class. At this point, these maps would not suffice to counter adversarial attacks. In fact, while it is true that they encode the essential frequency content required to predict a certain category, prior knowledge of the correct category would be required in order to choose the right map to apply. However, a simple modification in the map training strategy can address this issue.

Class-level essential frequency maps can be trained alternating the standard optimisation step, aimed at \textit{preserving} correct classification of clean data, with a second step aimed at \textit{reverting} adversarial attacks, akin to adversarial training \cite{goodfellow2015explaining}. Specifically, in the first step we consider a batch of clean images of category $c$ and aim for preserving the correct classification; in the second step we consider a batch of adversarially perturbed images, all misclassified in category $c$, and train the map to revert the classification to the original category of each image.

\looseness=-1 At test time, when an unseen and potentially adversarial sample is fed to the network, the initial prediction can be used to pick the corresponding map, so that a second forward pass can be executed on the filtered image, to produce the final prediction. This step proves helpful in counteracting adversarial attacks (Table \ref{attack_success}), though it imposes a toll in terms of clean accuracy, as in the case of most adversarial defence techniques \cite{raghunathan2020understanding}.

\looseness=-1 Moreover, as reported in Table \ref{attack_success}, in most cases these adversarially-trained class-level essential frequency maps can generalise across different adversarial attack algorithms, meaning that exact knowledge of the type of threat one might be facing is not a crucial requirement. While these findings alone are insufficient to build an end-to-end adversarial defence method, as an attacker may get access to the maps and produce new adversarial perturbations that deceive the whole model, we believe they represent a promising starting point for future research in this direction.

\begin{table}[h]
\centering

\caption{Clean and adversarial test accuracy when class-level essential frequency maps are applied to ResNet-20 for CIFAR-10 classification. Rows report different adversarial attacks (including no attack, i.e. clean accuracy), while on the columns we report the results relative to maps trained on different attacks, including the base case where no map is applied.}
\label{attack_success}
\centering
\begin{tabular}{ccccc}
    \toprule
 Data & \multicolumn{4}{c}{Map} \\ \cmidrule{2-5}
 &{\textbf{None}} & {\textbf{FMN}} & {\textbf{PGD}} &{\textbf{DeepFool}}\\
    \midrule
    \textbf{Clean} & 0.92 & 0.83 & 0.74 & 0.82\\
    \textbf{FMN}  & 0.04 & 0.78 & 0.72 & 0.78\\
    \textbf{PGD}  & 0.00 & 0.43 & 0.54 & 0.44\\
    \textbf{DeepFool} & 0.00 & 0.77 & 0.71 & 0.78\\
    \bottomrule
\end{tabular}
\end{table}

\section{Discussion}
Our study investigates the relationship between adversarial attacks and the implicit bias of neural networks in the Fourier domain. We focus on standard network architectures like ResNets and ViTs and datasets such as CIFAR-10 and Imagenette, where direct mathematical analysis of implicit bias is particularly challenging.
For the first time, we compute the essential frequencies required for an adversarial image to be misclassified. We demonstrate that these frequencies are highly correlated with the essential frequencies needed for a clean image to be correctly classified, using multiple correlation metrics.
Our work represents an advancement in understanding the nature of adversarial attacks and, more in general, the way the network represents the input. Apart from its impact on interpretability, identifying the crucial frequencies utilised by attackers holds promising implications for developing new defence and detection algorithms, thereby enhancing the security and robustness of neural networks. Notably, by using a variant of the essential frequency maps to revert the effects of adversarially attacked images, we show a new direction for crafting adversarial defences.

While studying the implicit bias of networks through the Fourier lens is a promising approach, that has been the focus of many recent works, it only offers a specific angle of analysis, which is a limitation of our study.
However, our map-based approach provides a flexible method for future investigations. For example, beyond examining the real part, researchers can investigate the essential Fourier phases in input images, opening new avenues for studying the implicit frequency bias of networks and their robustness. Additionally, other types of representations, such as wavelets, could be explored.
Another limitation of our study is that it focuses on the input frequencies neglecting those of all the intermediate representations within the network. Moreover, our approach can be extended to analyse the correlation for a broader class of frequency representations, for instance coming from different networks or data distributions. These areas of research will be the subject of future investigations.

In conclusion, our framework not only enhances the understanding of the way networks represent the input, with a particular focus on adversarially perturbed data, but it also paves the way for developing novel defence strategies, which could contribute to the overall advancement of AI security.
\section*{Acknowledgements}
We acknowledge AREA Science Park for making the supercomputing platform ORFEO available
for this work. Luca Bortolussi was supported by the PNRR project iNEST (Interconnected Nord-Est Innovation Ecosystem) funded by the European Union NextGenerationEU (Piano Nazionale di Ripresa e Resilienza
(PNRR) – Missione 4 Componente 2, Investimento 1.5 – D.D. 1058 23/06/2022, ECS 00000043,
CUP J43C22000320006).
\bibliographystyle{IEEEtran}
\bibliography{IEEEabrv,references}
\appendices

\section{Training details and hyperparameters}\label{train_details}
\looseness=-1 As introduced in Sec.~\ref{results}, we trained one model on CIFAR-10 \cite{krizhevsky2009learning} (ResNet-20 \cite{he2016deep}) and two on Imagenette \cite{imagenette}, namely a ResNet-18 \cite{he2016deep} and a ViT-B/16 \cite{dosovitskiy2020image}.
Here we provide additional details regarding the training (or fine-tuning) procedure of each model and the hyperparameters we chose.

ResNet-20 was trained from randomly initialised weights for $200$ epochs, using Stochastic Gradient Descent with momentum, set to $0.9$. The learning rate was initially set to $0.1$, to be decayed by a factor $10$ twice, after $100$ and $150$ epochs. $\ell_2$ regularisation was employed, by means of a weight decay factor of $10^{-4}$.
The final accuracy on the test set of CIFAR-10 was $92.29\%$.

For ResNet-18 and ViT we started from pre-trained ImageNet weights. Since Imagenette has $10$ classes, compared to the $1000$ of ImageNet, we had to replace the classification head, and we trained that layer only while keeping the other layers frozen. We trained the models for $20$ epochs using Adam \cite{kingma2014adam}, with a cyclic learning rate schedule, with maximum at $0.01$. We achieved final test accuracies of $98.32\%$ (ResNet-18) and $99.57\%$ (ViT).
\section{Details on the map training procedure}\label{map_training_app}
For all the modulatory maps we produced, we employed Adam optimiser \cite{kingma2014adam} with a learning rate of $0.01$. According to a criterion similar to early-stopping, maps were trained until convergence, and in any case for no less than $500$ optimisation steps each. We computed essential frequency maps only for correctly classified images,
while adversarial frequency maps were trained exclusively for correctly classified images that were successfully made adversarial by the attack. Sparsity of maps was enforced by means of $\ell_1$ regularisation, weighted by a factor $\lambda=0.5$ for CIFAR-10 and $\lambda=0.01$ for Imagenette.

The same setup was employed for training the class-level maps introduced in Sec.~\ref{robust_maps}, with the only difference that the training procedure was conducted on batches of images classified in each category instead of individual images. In such case, the level of sparsity to be enforced on the maps is much lower, as they need to encode a set of frequencies that covers multiple images together. Hence, $\lambda$ was set to $0.001$.
\end{document}